\documentclass[letterpaper, 10 pt, conference]{ieeeconf}  % Comment this line out if you need a4paper

\IEEEoverridecommandlockouts                              % This command is only needed if 
\usepackage{subcaption}
\usepackage{graphicx}
\usepackage{xcolor}
\usepackage{amsmath}
\usepackage{amssymb}
\usepackage{tabularx}
\usepackage{booktabs}
\usepackage{cleveref}
\usepackage{adjustbox}
\usepackage{todonotes}
\usepackage{footnote}
\usepackage{pifont}
\newcommand{\cmark}{\ding{51}}
\newcommand{\xmark}{\ding{55}}
\newcommand{\apex}{\textsc{apex} }

\newcommand{\bx}{\mathbf{x}} % image
\newcommand{\bm}{\mathbf{m}} % mask
\newcommand{\bz}{\mathbf{z}} % latents

\newcommand{\scene}{\mathcal{S}} % set of all objects making up the scene
\newcommand{\fg}{\mathcal{O}^{\text{fg}}} % set of fg objects
\newcommand{\bg}{\mathcal{O}^{\text{bg}}}
\newcommand{\setp}{\mathcal{O}^p}
\newcommand{\setd}{\mathcal{O}^d}
\newcommand{\zwhere}{\bz^{\text{where}}}
\newcommand{\zwhat}{\bz^{\text{what}}}
\newcommand{\zpres}{z^{\text{pres}}}
\newcommand{\zbg}{\bz^{\text{bg}}} % background object latents
\newcommand{\zfg}{\bz^{\text{fg}}} % foreground object latents
\newcommand{\zp}{\bz^{\text{p}}}
\newcommand{\zd}{\bz^{\text{d}}}

\title{\LARGE \bf
APEX: Unsupervised, Object-Centric Scene Segmentation and Tracking for Robot Manipulation
}

\author{Yizhe Wu$^{1}$, Oiwi Parker Jones$^{1}$, Martin Engelcke$^{1}$, and Ingmar Posner$^{1}$%
\thanks{$^{1}$Applied AI Lab, Oxford Robotics Institute, University of Oxford. Correspondence to: 
{\tt\small ywu@robots.ox.ac.uk}.}%
}

\begin{document}

\maketitle
\thispagestyle{empty}
\pagestyle{empty}

%%%%%%%%%%%%%%%%%%%%%%%%%%%%%%%%%%%%%%%%%%%%%%%%%%%%%%%%%%%%%%%%%%%%%%%%%%%%%%%%
\begin{abstract}
Recent advances in unsupervised learning for object detection, segmentation, and tracking hold significant promise for applications in robotics. A common approach is to frame these tasks as inference in probabilistic latent-variable models. In this paper, however, we show that the current state-of-the-art struggles with visually complex scenes such as typically encountered in robot manipulation tasks. We propose \textsc{apex}, a new latent-variable model which is able to segment and track objects in more realistic scenes featuring objects that vary widely in size and texture, including the robot arm itself. This is achieved by a principled mask normalisation algorithm and a high-resolution scene encoder.
To evaluate our approach, we present results on the real-world Sketchy dataset.
This dataset, however, does not contain ground truth masks and object IDs for a quantitative evaluation.
We thus introduce the Panda Pushing Dataset (P2D) which shows a Panda arm interacting with objects on a table in simulation and which includes ground-truth segmentation masks and object IDs for tracking.
In both cases, \textsc{apex} comprehensively outperforms the current state-of-the-art in unsupervised object segmentation and tracking.
We demonstrate the efficacy of our segmentations for robot skill execution on an object arrangement task, where we also achieve the best or comparable performance among all the baselines.
\end{abstract}

\section{Introduction}

% Motivation
Scene segmentation and tracking are cornerstones of robotics (e.g. \cite{Geiger2012CVPR,cordts2016cityscapes}).
A principal motivation is the ability to ground sensory observations in state-representations suitable for executing downstream tasks.
While considerable advances have been made in using supervised methods for detecting and segmenting objects (e.g. \cite{ren2015faster,he2017mask}), labelling data is resource intensive and quickly becomes intractable when trying to consider every possible object category for every deployment eventuality. 
Unsupervised learning -- the discovery of representations suitable for task execution without the need for training labels -- is therefore emerging as a promising alternative. 
In particular, recently developed \emph{object-centric} scene representations (e.g. \cite{burgess2019monet,crawford2019spatially,engelcke2020genesis,greff2019multi,jiang2020scalor,kosiorek2018sequential}) %,
have the potential of vastly improving data efficiency in robotics and other applications (e.g. \cite{veerapaneni2020entity,wulfmeier2020representation,watters2019cobra}). %
Here we show, however, that current state-of-the-art methods fail on visually complex datasets that are representative of scenarios commonly encountered in robot manipulation.

% Problem background
Within the class of unsupervised, object-centric models, variational autoencoders (VAEs) (\cite{kingma2013auto,rezende2014stochastic}) are emerging as a popular choice. Object detection and segmentation are performed by using spatial transformer networks (STNs) \cite{jaderberg2015spatial} and spatial Gaussian mixture models (SGMMs) (\cite{greff2016tagger,greff2017neural,van2018relational}) to separate objects, respectively.
Some works do not impose any further structure on these latent representations (e.g. \cite{burgess2019monet,engelcke2020genesis,veerapaneni2020entity,locatello2020object}), while others further factorise the representations to explicitly disentangle foreground objects from the background, object location, appearance, and whether an object slot is used or not (e.g. \cite{crawford2019spatially,jiang2020scalor,eslami2016attend,huang2015efficient}).%
In contrast to, e.g., \textsc{op3} \cite{veerapaneni2020entity} which was also developed in the context of robotics, we argue the latter is of particular utility in robotics applications where such highly structured representations can be directly used as inputs to a planning module.
A recently proposed model of this type called \textsc{scalor} \cite{jiang2020scalor} achieves particularly good results on segmenting and tracking objects in videos that contain a possibly large number of objects.
We demonstrate however, that \textsc{scalor} struggles to learn object-centric representations on datasets with objects of widely varying sizes and textures as encountered in robot manipulation.
\begin{figure*}
    \centering
    \begin{subfigure}{0.44\linewidth}
        \includegraphics[trim=0 93 0 0, clip, width=\linewidth]{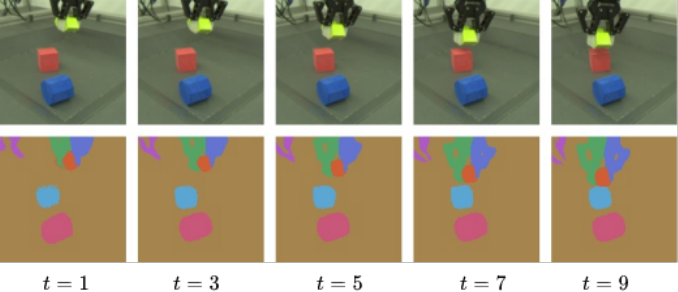}
        \vspace{-0.55cm}
        \caption{Sketchy}
    \end{subfigure}
    \qquad
    \begin{subfigure}{0.44\linewidth}
        \includegraphics[trim=0 3 0 85, clip, width=\linewidth]{figs/demo_recon_rp_sk_v2.pdf}
        \vspace{-0.65cm}
        \caption{Panda Pushing Dataset (P2D)}
    \end{subfigure}
    \vspace{-0.2cm}
    \caption{\textsc{apex} learns to segment and track objects in videos \emph{without supervision} from (a) the established Sketchy dataset~\cite{cabi2019scaling} and (b) our Panda Pushing Dataset (P2D). Segment colour indicates object ID. \textsc{apex} accurately segments the diverse objects in the scenes and tracks the blue cup in (b) despite severe occlusion at $t=\{7,11\}$.} 
    \label{fig:teaser_fig}
\end{figure*}

We therefore develop \textsc{apex} (Amortised Parallel infErence with miXture models), a novel object-centric generative model trained on videos.
In contrast to prior work, \textsc{apex} uses a principled mask normalisation procedure to parameterise an SGMM, which allows the explicit tuning of foreground and background standard deviations.
\textsc{apex} also features an improved scene encoder that outputs a high-resolution feature map that is particularly suitable for tracking \cite{zhou2019objects}.
To showcase the efficacy of \textsc{apex}, we evaluate \textsc{apex} qualitatively on Sketchy \cite{cabi2019scaling}, an existing real-world robot manipulation dataset.
We also introduce the Panda Pushing Dataset (P2D) as a quantitative benchmark against prior art.
This contains videos of a Panda arm interacting with objects on a table in simulation and includes ground truth labels for segmentation masks and tracking objects between frames.
It can be observed that \textsc{apex} comprehensively outperforms recent state-of-the-art methods (\cite{jiang2020scalor,veerapaneni2020entity,locatello2020object,lin2020space,lin2020improving}) by a large margin in terms of unsupervised segmentation and object tracking. 
Finally, inspired by the \textit{Open Cloud Robot Table Organization Challenge}~\cite{OCRTOC}, we demonstrate the utility of \apex specifically in the context of a robot object re-arrangement task.
Here, the superior segmentation performance of \textsc{apex}, as illustrated in \Cref{fig:teaser_fig}, as well as the quality of the learned object representations lead to significantly better results compared to prior art.

\section{Related Work}
This work builds on recent literature on object-centric generative models (OCGMs), which are typically formulated as VAEs (e.g. \cite{eslami2016attend,huang2015efficient}) or generative adversarial networks (GANs) (e.g. \cite{ehrhardt2020relate,nguyen2020blockgan,niemeyer2020giraffe}).%van2018case
Unlike object-centric GANs, VAE-based methods directly provide an amortised inference mechanism for extracting object-centric representations from input images.
Early works that use STNs for separating objects do so by sequentially attending to different regions in an image, leading to a computational complexity that increases linearly with the number of objects (\cite{eslami2016attend,huang2015efficient}).
More recent works parallelise the inference of object representations, which has been shown to be particularly useful for images with a large number of objects (\cite{crawford2019spatially,jiang2020scalor}).
Segmentation-based models that parameterise an SGMM (e.g. \cite{burgess2019monet,engelcke2020genesis,greff2019multi,locatello2020object}) tend to be computationally more expensive than STN-based approaches where objects can be generated as smaller crops rather than image-sized components.
A more informative pixel-level labelling, however, is required in applications such as the object arrangement task considered in this work.
\textsc{slot-attention} \cite{locatello2020object} is a prominent recent model belonging to this category and is therefore selected as one of our baselines.
The majority of related works learn object representations from individual images (e.g. \cite{burgess2019monet,engelcke2020genesis,locatello2020object}), but some also exploit temporal information in video sequences (e.g. \cite{jiang2020scalor,kosiorek2018sequential,veerapaneni2020entity}) to improve object separation.
\apex also leverages a VAE, STNs, and an SGMM for the parallel inference of structured, object-centric latent representations from videos.
The structure of the model is most closely related to \textsc{scalor} \cite{jiang2020scalor}.
Instead of parameterising a Gaussian image likelihood, however, \apex parameterises an SGMM which allows direct tuning of loss magnitudes from background and foreground modules.
Moreover, \apex uses a scene encoder that is particularly suitable for object tracking \cite{zhou2020tracking}.

A small number of works also explore the use of object-centric representations in robotics (\cite{veerapaneni2020entity,wulfmeier2020representation}).
\textsc{op3} \cite{veerapaneni2020entity} uses an object-centric model to predict goal images that contain a set of blocks in a desired configuration and the authors in \cite{wulfmeier2020representation} show that explicit object representations can accelerate the acquisition of 
robotic manipulation skills.
Similarly, \textsc{cobra} \cite{watters2019cobra} leverages object-centric representations to improve data efficiency and policy robustness in several RL tasks in visually simple simulated environments.
Inspired by \cite{OCRTOC}, we benchmark a number of models on an object manipulation task, where we observe that segmentations and object representations learned by \textsc{apex} lead to significantly better results compared to the recent state-of-the-art.

%%%%%%%%%%%%%%%%%%%%%%%%%%%%%%%%%%%%%%%%%%%%%%%%%%%%%%%%%%%%%%%%%%%%%%%%%%%%%%%%
\section{APEX: Amortised Parallel Inference with Mixture Models}
\label{section:apex}
%%%%%%%%%%%%%%%%%%%%%%%%%%%%%%%%%%%%%%%%%%%%%%%%%%%%%%%%%%%%%%%%%%%%%%%%%%%%%%%%

Let $\bx \in [0, 1]^{H \times W \times C}$ be a frame from a video sequence $\bx_{1:T}$, where $H$ and $W$ denote the image height and width, $C$ represents the number of image channels (e.g. RGB), and $T$ denotes the number of frames in the video sequence.
Consider a scene $\scene$ to be formed of $K$ object hypotheses -- or \emph{components} -- which are each encoded by a set of latent variables such that $\scene = \{\bz_1, \dots,\bz_{K}\}$.
In particular, we consider two disjoint sets of object hypotheses comprised of \emph {foreground} objects and a \emph{background} component, such that $\scene = \fg \cup \bg$, where $\fg = \{\zfg_1, \dots , \zfg_{K-1}\}$ and $\bg = \{\zbg\}$, i.e., $\zbg = \bz_K$.
Each foreground component is described by a set of latent variables $\zfg_k = \{\zwhere_k, \zwhat_k, \zpres_k\}$ encoding its location, appearance, and existence in the scene (see \cite{eslami2016attend}).
$\zbg$ directly encodes the appearance of the background.
For each frame $\bx_t$ at time $t$, foreground components can either be \emph{propagated} from the previous time step %$t-1$ 
or \emph{discovered} in the current image (see \cite{jiang2020scalor,kosiorek2018sequential}).

\apex defines a generative model with learnable parameters $\theta$ that is formulated as an SGMM (e.g. \cite{burgess2019monet,engelcke2020genesis,greff2019multi}) via the image likelihood
\begin{equation}
p_{\theta}(\bx_t|\bz_{1:K,t}) = \sum_{k=1}^K \bm_{k}(\bz_{1:K,t}) \odot \mathcal{N}(\mu(\bz_{k,t}),\sigma_k),
\label{eq:pixel_likelihood}
\end{equation}
where $\bm_{k,t}$ are the segmentation masks, $\mu_{k,t}$ are the means of the Gaussian components, and $\sigma_k$ is a fixed component standard deviation.
Separate standard deviations are used for the foreground and background, i.e. $\sigma_{1:K-1} = \sigma^{\mathrm{fg}}$ and $\sigma_{K} = \sigma^{\mathrm{bg}}$.
A probabilistic prior $p_\theta(\bz_{1:K,1:T})$ is defined to regularise the model.

\begin{figure*}
    \centering
    \includegraphics[width=0.9\linewidth]{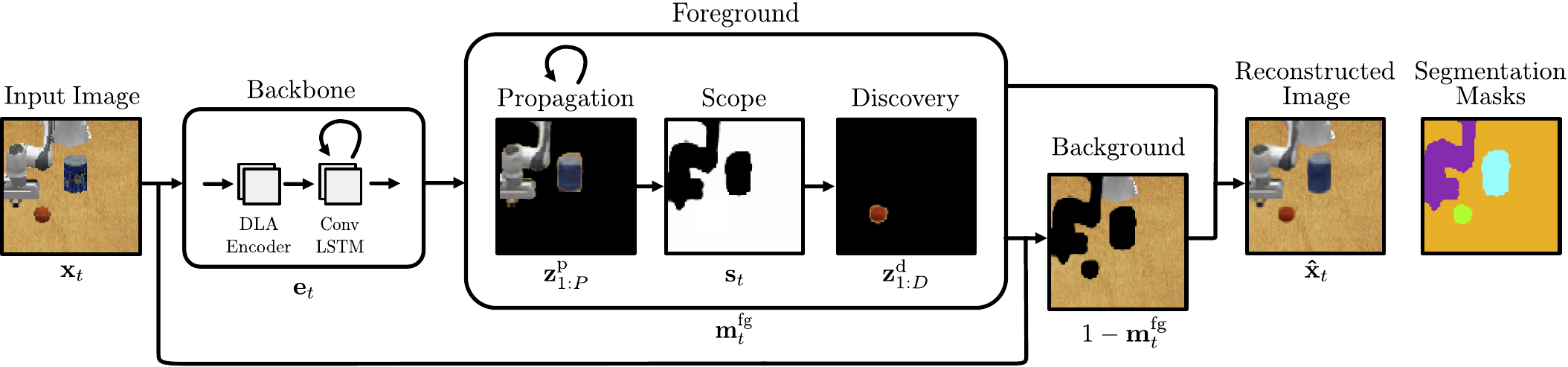}
    \vspace{-0.1cm}
    \caption{Illustration of \textsc{apex}. A backbone first extracts an encoding from an input image. This is used to propagate objects from the previous time step and to discover additional objects in the current image. The remainder of the image is treated as background. The final image is composed from both the foreground objects and the background.}
    \label{fig:apex}
\end{figure*}

To achieve segmentation and tracking, \apex provides an approximate inference model \cite{jordan1999introduction} with learnable parameters $\phi$ for an object-centric latent representation of how a scene evolves in a sequence of images, i.e.,
\begin{equation}
q_\phi(\bz_{1:K,1:T}|\bx_{1:T}) = \prod_{t=1}^{T} q_\phi(\bz_{1:K,t}|\bz_{1:K,<t}, \bx_{\leq t}) \,.
\end{equation}
The inference and generative models are learned jointly as a VAE (\cite{kingma2013auto,rezende2014stochastic}).
An overview of \apex is illustrated in \Cref{fig:apex}.
This section proceeds by first defining the underlying structure of the generative model, before defining the associated inference model that facilitates the segmentation and tracking of objects in video sequences.

%%%%%%%%%%%%%%%%%%%%%%%%%%%%%%%%%%%%%%%%%%%%%%%%%%%%%%%%%%%%%%%%%%%%%%%%%%%%%%%%
\subsection{Generative Model}
To capture correlations between frames, we assume the latents $\bz_{1:K,t}$ depend on the latents of the previous time steps
\begin{equation}
p_\theta(\bz_{1:K,1:T}) = \prod_{t=1}^{T} p_\theta(\bz_{1:K,t}|\bz_{1:K,<t}) \,.
\label{eq:gen_video}
\end{equation}
$p_\theta(\bz_{1:K,t})$ is further factorised into three terms to represent \emph{propagated} objects $\setp_t$, \emph{discovered} objects $\setd_t$, and the background $\bg_t$.
Assuming there are $P$ propagated objects and $D$ newly discovered objects, we assume propagated objects depend on the previous latent variables, newly discovered objects in turn depend on objects that have already been propagated to avoid the \emph{rediscovery} of propagated objects
\begin{multline}
p_\theta(\bz_{1:K,t}|\bz_{1:K,<t}) = \\
p_\theta(\zbg_{t}| \zbg_{<t}) \, p_\theta(\zd_{1:D,t}|\zp_{1:P,t}) \, p_\theta(\zp_{1:P,t} | \bz_{1:K,<t}) \, .
\label{eq:latent_factor_t}
\end{multline}
To facilitate application in robotics and similar to prior works (e.g. \cite{jiang2020scalor,kosiorek2018sequential}), the latents describing foreground objects in $\fg$ are factorised into $\{\zwhere_k, \zwhat_k, \zpres_k\}$ describing component location, appearance, and presence in the scene, respectively, such that
\begin{multline}
p_\theta(\zp_{k,t}|\zfg_{k,<t}) = \\
p_\theta(\zpres_{k,t}|\zfg_{k,<t}) \, p_\theta(\zwhat_{k,t}|\zfg_{k,<t})^{\zpres_{k,t}} \,
p_\theta(\zwhere_{k,t}|\zfg_{k,<t})^{\zpres_{k,t}} \,,
\end{multline}
\begin{multline}
p_\theta(\zd_{k,t}|\zp_{1:P,t}) = \\
p_\theta(\zpres_{k,t}|\zp_{1:P,t}) \, p_\theta(\zwhat_{k,t})^{\zpres_{k,t}} \,
p_\theta(\zwhere_{k,t})^{\zpres_{k,t}} \,.
\end{multline}

The segmentation masks $\bm_{k}$ and the means $\mu$ of the Gaussian components in \Cref{eq:pixel_likelihood} are decoded from $\{\zwhere_{k,t},\zwhat_{k,t},\zpres_{k,t}\}$ and $\zbg_t$. We consider $\zpres_{k,t}$ to be Bernoulli distributed and the remainder of the latents being Gaussian distributed (see \cite{jiang2020scalor,lin2020space}). $\zwhat_{k,t}$ encodes the appearance of a component and the logits $\alpha_{k,t}$ from which the mask $\bm_{k,t}$ is later obtained.
$\zbg_t$ only encodes the appearance of the background component.
$\zwhere_{k,t}$ encodes a transformation in an STN \cite{jaderberg2015spatial}, describing the size and location of a bounding box that contains an object.
A separate variable for encoding background location is not needed as every pixel that is not assigned to a foreground object is treated as background.
%$\bm^{\text{fg}}_t$

In contrast to \textsc{scalor}, where the foreground mask needs to be  explicitly clamped to $\left[0,1\right]$, we employ a principled approach to mask normalisation. In particular, we first introduce a foreground mask $\bm^{\text{fg}}_t$ to model the occupancy of the foreground as a whole, and then compute intermediate object masks $\hat{\bm}_{1:K-1,t}$ which attribute specific occupancy responsibility to each individual object. 
Specifically, the mask logits $\alpha_{1:K-1,t}$ for the $K-1$ foreground objects are computed as
\begin{equation}
\alpha_{k,t} =c\, \operatorname{tanh}\left( \mathrm{CNN}(\zwhat_{k,t}) \right) \,,
\label{eq:alpha}
\end{equation}
where $c$ is a fixed constant that constrains the mask logits to $[-c\,, c]$.
The foreground mask $\bm^{\text{fg}}_t$ is computed as
\begin{equation}
\bm^{\text{fg}}_t = \operatorname{tanh}\left(\sum _{k=1}^{K-1}\operatorname{STN}\left( \operatorname{softplus}(\alpha_{k,t}) \, \zpres_{k,t},\zwhere_{k,t}\right) \right)  \,,
\label{eq:mask_occupied}
\end{equation}
where the softplus operation ensures that a possible foreground component makes a non-negative contribution to the foreground mask when $\zpres_{k,t} = 1$.
The intermediate object masks $\hat{\bm}_{1:K-1,t}$ are obtained such that
\begin{align}
\hat{\alpha}_{k,t} &= \operatorname{STN}\left(\alpha_{k,t},\zwhere_{k,t}\right) + 2c\zpres_{k,t} \,, \label{eq:alpha_hat}\\
\hat{\bm}_{1:K-1,t} &= \operatorname{softmax} \left(\hat{\alpha}_{1,t}, \, \dots \, , \, \hat{\alpha}_{K-1,t}\right) \,.
\label{eq:m_hat}
\end{align}
\Cref{eq:alpha_hat} maps components with $\zpres_{k,t} = 1$ to an interval that does not overlap with components where $\zpres_{k,t} = 0$. This formalises the intuition that objects which are \emph{not present} will not occupy any physical space.
The masks for the $K-1$ foreground objects are then obtained by the element-wise multiplication of the foreground mask $\bm^{\text{fg}}_t$ with intermediate object masks $\hat{\bm}_{1:K-1,t}$,
\begin{equation}
    \bm_{1:K-1,t} = \hat{\bm}_{1:K-1,t} \odot \bm^{\text{fg}}_t \,.
\end{equation}
Finally, inspired by the stick-breaking process formulations in prior work (\cite{burgess2019monet,engelcke2020genesis}), the background mask $\bm^{\text{bg}}_{t}$ is computed as
\begin{equation}
    \bm^{\text{bg}}_{t} = 1 - \bm^{\text{fg}}_t \,.
\end{equation}

%%%%%%%%%%%%%%%%%%%%%%%%%%%%%%%%%%%%%%%%%%%%%%%%%%%%%%%%%%%%%%%%%%%%%%%%%%%%%%%%
\subsection{Inference Model}
The true posterior over latent variables is generally intractable, so a variational approximation $q_\phi(\bz_{1:K,1:T}|\bx_{1:T})$ is introduced.
Mirroring \Cref{eq:gen_video} in the generative model, the approximate posterior is factorised as
\begin{equation}
q_\phi(\bz_{1:K,1:T}|\bx_{1:T}) = \prod_{t=1}^{T} q_\phi(\bz_{1:K,t}|\bz_{1:K,<t}, \bx_{\leq t}) \,.
\end{equation}
The posterior at time step $t$ exhibits the same dependencies as were assumed in \Cref{eq:latent_factor_t} and can thus be written as
\begin{multline}
q_\phi(\bz_{1:K,t}|\bz_{1:K,<t}, \bx_{\leq t}) = q_\phi(\zbg_{t}| \zd_{1:D,t},\zp_{1:P,t},\bx_{t}) \\
q_\phi(\zd_{1:D,t} | \zp_{1:P,t}, \bx_{\leq t}) \,
q_\phi(\zp_{1:P,t} | \bz_{1:K,<t}, \bx_{\leq t}) \,.
\label{eq:posterior_t}
\end{multline}
We assume that the approximate posterior for the $k^{\text{th}}$ propagated object in $\setp_t$ factorises such that
\begin{align}
\begin{split}
& q_\phi(\zp_{k,t}|\zfg_{k,<t},\bx_{\leq t}) = \\
& \qquad q_\phi(\zpres_{k,t}|\zwhat_{k,\leq t},\zpres_{k,<t},\zwhere_{k,t-1},\bx_{\leq t}) \\
& \qquad q_\phi(\zwhat_{k,t}|\zwhere_{k,t},\zwhat_{k,<t},\bx_{\leq t})^{\zpres_{k,t}} \\
& \qquad q_\phi(\zwhere_{k,t}|\zwhere_{k,<t},\zwhat_{k,<t},\bx_{\leq t})^{\zpres_{k,t}} \,
\end{split}
\end{align}
and that the posterior for a discovered object factorises as
\begin{multline}
q_\phi(\zd_{k,t}|\zp_{1:P,t},\bx_{\leq t}) = \,
q_\phi(\zpres_{k,t}|\zwhat_{k,t},\zwhere_{k,t},\zp_{1:P,t},\bx_{\leq t}) \\
q_\phi(\zwhat_{k,t}|\zwhere_{k,t},\bx_{\leq t})^{\zpres_{k,t}} \,
q_\phi(\zwhere_{k,t}|\bx_{\leq t})^{\zpres_{k,t}} \,.
\end{multline}
Intuitively, $\zwhere_{k,t}$ encodes where to look in an image in order to infer $\zwhat_{k,t}$ which is used for determining the appearance of an object.

%%%%%%%%%%%%%%%%%%%%%%%%%%%%%%%%%%%%%%%%%%%%%%%%%%%%%%%%%%%%%%%%%%%%%%%%%%%%%%%%
\subsection{Inference - Implementation}

We now proceed with describing the implementation of how these posterior distributions are inferred for object \emph{propagation} and \emph{discovery} as well as the \emph{background}.

%%%%%%%%%%%%%%
\noindent
\textbf{Feature Extraction}\ \
Features are extracted with a shared encoder.
Observations $\bx_{\leq t}$ are stacked and encoded into a feature map $\mathbf{e}_t \in \mathbb{R}^{H/4 \times W/4 \times F}$ with a deep layer aggregation (DLA) encoder~\cite{yu2018deep} and a ConvLSTM~\cite{xingjian2015convolutional} to capture spatio-temporal correlations.
$F$ is the number of the feature channels.
The encoder outputs a high-resolution feature map that preserves spatial correspondences between the features and input images.
This is beneficial for object tracking \cite{zhou2019objects} and in contrast to \textsc{scalor}, where a low resolution feature map is used instead.
\textit{Propagation:}\ \
Using the previous object bounding box at $t-1$, a square feature map $\mathbf{f}^{\text{p}}_{k,t}$ is extracted
from $\mathbf{e}_t$ using an STN, whereby the square dimensions are set equal to the larger side of the previous bounding box.
A CNN then maps $\mathbf{f}^{\text{p}}_{k,t}$ to a feature vector $\mathbf{c}^{\text{p}}_{k,t}$.
\textit{Discovery:}\ \
A grid of equally spaced features $\mathbf{c}^{\text{d}}_{1:D,t}$ is extracted from the feature map $\mathbf{e}_t$ to discover new objects.

%%%%%%%%%%%%%%
\noindent\textbf{Object Location}\ \
\textit{Propagation:}\ \
The posterior over $\bz^{where}_{k,t}$ is computed with an RNN whose inputs are $[\mathbf{c}^{\text{p}}_{k,t},\, \zwhat_{k,t-1},\, \zwhere_{k,t-1}]$.
\textit{Discovery:}\ \
The posterior over $\zwhere_{k,t}$ is computed from $\mathbf{c}^{\text{d}}_{k,t}$ with a $1 \times 1$ convolution.

%%%%%%%%%%%%%%
\noindent\textbf{Object Appearance}\ \
\textit{Propagation:}\ \
A \emph{glimpse} $\mathbf{G}_{k,t}$ is cropped from the feature map $\mathbf{e}_t$ using a STN according to $\zwhere_{k,t}$.
A RNN takes the encoding of the glimpse $\mathbf{G}_{k,t}$ encoded by a CNN and $\zwhat_{k,t-1}$ as inputs to infer the posterior over $\zwhat_{k,t}$. 
\textit{Discovery:}\ \
The posterior over $\zwhat_{k,t}$ is inferred in same fashion as for propagated objects with weights being shared and $\zwhat_{0}$ being initialised to a vector of zeros.
\textit{Background:}\ \
The posterior over $\zbg_{t}$ is obtained with a CNN using the background mask $\bm^{\text{bg}}_t$ and the current image $\bx_t$.

%%%%%%%%%%%%%%
\noindent\textbf{Object Presence}\ \
\textit{Propagation:}\ \
The posterior over $\zpres_{k,t}$ is computed with an RNN whose inputs are $[\mathbf{c}^{\text{p}}_{k,t},\, \zwhat_{k,t},\, \zpres_{k,t-1}]$.
\textit{Discovery:}\ \
The presence of objects in the \textit{discovery} phase is determined by: 1. whether a new object is detected and 2. whether that object has been explained by the propagated objects.
We use the scope $\mathbf{s}_t$ to indicate which pixels have not been explained yet.
This scope is defined as $\mathbf{s}_t = 1 - \bm^\text{p}_t$ whereby $\bm^\text{p}_t$ is computed as in \Cref{eq:mask_occupied} but only using propagated objects.
We thus decompose the posterior over $\zpres_{k,t}$ into two terms $\{p^{\text{proposal}}_{k,t}, p^{\text{context}}_{k,t}\} \in [0,1]$ so that
\begin{equation}
    q_\phi(\zpres_{k,t}|\zwhat_{k,t},\zwhere_{k,t},\zp_{1:P,t},\bx_{\leq t}) = p^{\text{proposal}}_{k,t} \, p^{\text{context}}_{k,t} \,.
\end{equation}
$p^{\text{proposal}}_{k,t}$ is obtained with a fully-connected layer from $[\mathbf{c}^{\text{d}}_{k,t},\, \zwhat_{k,t},\, \zwhere_{k,t}]$ and describes whether a cell might contain a new object or not.
The purpose of $p^{\text{context}}_{k,t}$ is to avoid the rediscovery of objects via the scope $\mathbf{s}_t$.
This removes discovered objects that overlap with propagated objects and it is computed as 
\begin{equation}
    \mathbf{p}^{\text{context}}_{k,t} = 
    \frac{ \sum_{i,j} \mathbf{s}_{t,i,j} \, \operatorname{tanh}(\operatorname{softplus}(\alpha_{k,t,i,j}))} {\sum_{i,j} \operatorname{tanh}(\operatorname{softplus}(\alpha_{k,t,i,j}))} \,,
\end{equation}
where $(i,j)$ are all pixel coordinate tuples in an image. Intuitively, $\mathbf{p}^{\text{context}}_{k,t}$ corresponds to the fraction of the proposed object mask that has not been explained by the propagated objects.
\textit{Object Filtering:}\ \
To reduce memory requirements, foreground objects are discarded at every time step when $\zpres_{k,t}$ is below a fixed, manually set threshold.

%%%%%%%%%%%%%%%%%%%%%%%%%%%%%%%%%%%%%%%%%%%%%%%%%%%%%%%%%%%%%%%%%%%%%%%%%%%%%%%%
\subsection{Learning}
The inference and generative models can be jointly trained by maximising the evidence lower bound (ELBO).
Omitting object subscripts, this is given by 
\begin{align}
\mathcal{L}(\mathbf{\theta},\mathbf{\phi}) =& \sum_{t=1}^{T} \mathbb{E}_{q_{\mathbf{\phi}}(\bz_{t}|\bz_{<t},\bx_{\leq t})} \left[\log p_{\theta}(\bx_{t}|\bz_{t})\right]\\
&+ \mathbb{KL}\left[q_{\mathbf{\phi}}(\bz_t|\bz_{<t},\bx_{\leq t})\parallel p_{\mathbf{\theta}}(\bz_t|\bz_{<t})\right] \,.
\end{align}
Prior distributions for continuous and discrete variables are assumed to be Gaussian and Bernoulli, respectively.
Continuous variables are reparameterised (see \cite{kingma2013auto,rezende2014stochastic}) and discrete variables are obtained using the Gumbel-Softmax trick \cite{jang2016categorical}. In practice, we minimise the inclusive KL divergence~\cite{li2017dropout} for the $\zwhat$ and $\zwhere$ as we empirically find this leads to better performance. 
We also include an entropy loss on the $K-1$ foreground object masks in the training objective,
\begin{equation}
\mathcal{L}_{H} = \sum^H_{i=1} \sum^W_{j=1} \sum^T_{t=1} \sum^{K-1}_{k=1} -m^{\text{fg}}_{t,i,j} \, m_{k,t,i,j} \, \log m_{k,t,i,j}\,  \,.
\end{equation}
This penalises pixels being explained by multiple components. The full training objective is the sum of the ELBO and the mask entropy loss:
\begin{equation}
\mathcal{L}=\mathcal{L}(\mathbf{\theta},\mathbf{\phi}) + \mathcal{L}_{H} \,.
\end{equation}

\section{Experiments}
\label{section:experiments}
This section presents experiments on unsupervised scene segmentation, object tracking, and a simulated object manipulation task to showcase the capabilities of \textsc{apex}.
Our recent, state-of-the-art baselines consist of \textsc{scalor} \cite{jiang2020scalor}, \textsc{op3} \cite{veerapaneni2020entity}, \textsc{slot-attention} \cite{locatello2020object}, \textsc{space} \cite{lin2020space}, and \textsc{g-swm} \cite{lin2020improving}.

%%%%%%%%%%%%%%%%%%%%%%%%%%%%%%%%%%%%%%%%%%%%%%%%%%%%%%%%%%%%%%%%%%%%%%%%%%%%%%%%
{\raggedleft{\bf Datasets}}
We perform a qualitative evaluation using the real-world Sketchy dataset \cite{cabi2019scaling}, which contains demonstration trajectories of a robot arm performing different tasks involving a set of objects.
The images are pre-processed as in ~\cite{engelcke2021genesisv2}, using a $128 \times 128$ resolution and sequences of length 10.
Sketchy, however, does not contain object annotations, which prohibits the quantitative evaluation of object segmentation and tracking methods.
We therefore introduce the Panda Pushing Dataset (P2D), which shows a Panda arm interacting with objects in simulation and includes pixel-level ground truth segmentations as well as object tracking IDs.
Up to three objects are spawned in each episode and the robot-arm moves along a randomly selected straight line in the horizontal direction.
Objects in the dataset are sampled from a set of 14 common objects (e.g. mugs, coffee cans, apples) with varying shapes, colours, and textures (see \cite{calli2017yale}).
We collect a total of 2,400 trajectories (2,000 for training, 200 for validation, and 200 for testing) with each trajectory having a length of 20 frames with a resolution of $128 \times 128$.

%%%%%%%%%%%%%%%%%%%%%%%%%%%%%%%%%%%%%%%%%%%%%%%%%%%%%%%%%%%%%%%%%%%%%%%%%%%%%%%%
{\raggedleft{\bf Metrics}}
Similar to prior work (e.g.~\cite{engelcke2020genesis,greff2019multi}), the quality of object segmentations is evaluated using the Adjusted Rand Index (ARI)~\cite{rand1971objective} and the Mean Segmentation Covering (MSC) on a held-out test set.
To enable rigorous benchmarking, we report results on two variants of these metrics: one which only considers foreground objects (see \cite{engelcke2020genesis}) as well as one where all pixels and ground truth masks are considered, including those belonging to the background.
The latter is relevant in the context of this work as the background needs to be explicitly separated from the foreground for the manipulation task in \Cref{sec:arrangement}.
Multi-object tracking metrics are evaluated following the procedure in Weis et al.~\cite{Weis2020}, which is based on the protocol from the established MOT16 tracking benchmark \cite{milan2016mot16}.

%%%%%%%%%%%%%%%%%%%%%%%%%%%%%%%%%%%%%%%%%%%%%%%%%%%%%%%%%%%%%%%%%%%%%%%%%%%%%%%%
{\raggedleft{\bf Implementation Details}}
\textsc{apex} is trained with ADAM \cite{kingma2014adam} and a learning rate of $10^{-4}$.
The baselines are trained with the default learning rates and optimisers from the released implementations.
The number of components in \textsc{op3} and \textsc{slot-attention} is set to $K=5$ for P2D %(up to three objects, the arm, and the background) 
and to $K=8$ for Sketchy. %(up to three objects, the arm base, the left and right gripper, the cable and the background).
For \textsc{scalor}, the hard constraint on object size as found in the original implementation is removed as P2D and Sketchy contain objects of various sizes.
\textsc{apex}, \textsc{scalor}, and \textsc{op3} are trained with a batch size of $4$ for $4 \times 10^4$ iterations.
An additional $10^4$ warm-up iterations are used for \textsc{g-swm}. %during which the sequence length as incrementally increased, similar as in the original work
With the exception of \textsc{op3}, the models are trained on the full image sequences.
\textsc{op3} is trained on sub-sequences of length five due to the model capacity limitations on images that are resized to $64\times64$ as used in the original model.
Training \textsc{apex}, \textsc{scalor}, \textsc{g-swm} and \textsc{op3} on a single NVIDIA Titan RTX GPU takes about 20 hours each.
For \textsc{space}, the batch size is increased to $16$ and the number of training iterations are increased to $2 \times 10^5$ and $1 \times 10^5$ iterations for P2D and Sketchy, respectively, to account for the fact that \textsc{space} is trained on individual images rather than image sequences.
For \textsc{slot-attention} the number of training iterations is further increased to $4 \times 10^5$ and $2.5 \times 10^5$ iterations for P2D and Sketchy, respectively, as we found that the models take longer to learn reasonable segmentation masks.
The resulting wall-clock times for \textsc{space} and \textsc{slot-attention} are about 5 hours and 20 hours, respectively.

%%%%%%%%%%%%%%%%%%%%%%%%%%%%%%%%%%%%%%%%%%%%%%%%%%%%%%%%%%%%%%%%%%%%%%%%%%%%%%%%
\subsection{Unsupervised Segmentation and Tracking}

Quantitative results for unsupervised segmentation and tracking on P2D are summarised in \Cref{table:segmentation_metrics} and \Cref{table:tracking_metrics}, respectively.
\begin{table*}
    \centering
    \caption{Mean and standard deviation of the segmentation metrics on P2D from four random seeds.}
    \label{table:segmentation_metrics}
    \begin{tabular}{l c c c c c}
        \toprule
        & Object tracking & ARI-FG & MSC-FG & ARI & MSC \\
        \midrule
        \textsc{op3} & \cmark & $0.30\pm0.03$ & $0.30\pm0.01$ & $0.15\pm0.05$ & $0.39\pm0.03$  \\
        \textsc{scalor} & \cmark & $0.31\pm0.03$ & $0.37\pm0.01$ & $0.43\pm0.04$ & $0.52\pm0.01$ \\
        \textsc{g-swm} * & \cmark & $0.38\pm0.17$ & $0.42\pm0.05$ & $0.33\pm0.13$ & $0.55\pm0.04$ \\
        \textsc{space} & \xmark & $0.33\pm0.17$ & $0.49\pm0.12$ & $0.72\pm0.21$ & $0.62\pm0.09$ \\
        \textsc{slot-att.} & \xmark & $0.77\pm0.24$ & $0.46\pm0.16$ & $0.25\pm0.22$ & $0.50\pm0.14$ \\
        \apex & \cmark & $\mathbf{0.89\pm0.02}$ & $\mathbf{0.73\pm0.01}$ & $\mathbf{0.93\pm0.00}$ & $\mathbf{0.80\pm0.01}$ \\
        \bottomrule
        \multicolumn{6}{l}{* The \textsc{g-swm} results are computed with one failed random seed being excluded.}\\
    \end{tabular}
\end{table*}
\begin{table*}
    \centering
    \caption{Mean and standard deviation of the tracking metrics on P2D from four random seeds.}
    \label{table:tracking_metrics}
    \begin{adjustbox}{max width=1.0\textwidth}
    \begin{tabular}{l c c c c c c c c}
        \toprule
        & MOTA $\uparrow$ & MOTP $\uparrow$ & Match $\uparrow$ & ID S. $\downarrow$ & FPs $\downarrow$ & Miss $\downarrow$ & MD $\uparrow$ & MT $\uparrow$ \\
        \midrule
        \textsc{op3} & $-48.6\pm17.0$ & $66.6\pm0.9$ & $25.7\pm4.4$ & $0.3\pm0.1$ & $74.4\pm13.4$ & $73.9\pm4.5$ & $15.2\pm4.7$ &$15.6\pm4.8$\\
        \textsc{scalor} & $-125.6\pm6.3$ & $73.5\pm0.5$ & $31.6\pm1.1$ & $2.6\pm0.0$ &$157.2\pm6.5$ &$65.8\pm1.1$ &$19.0\pm1.7$ &$22.3\pm2.3$\\
        \textsc{g-swm} * & $-41.8\pm2.0$ & $73.1\pm0.5$ & $36.7\pm4.0$ & $2.0\pm1.0$ &$78.4\pm2.5$ &$61.3\pm3.5$ &$25.3\pm4.6$ &$28.6\pm3.2$\\
        \apex & $\mathbf{50.5\pm4.0}$ & $\mathbf{83.2\pm0.6}$ & $\mathbf{79.7\pm0.5}$ & $\mathbf{0.2\pm0.1}$ & $\mathbf{29.2\pm3.8}$ & $\mathbf{20.0\pm0.5}$ & $\mathbf{70.8\pm1.0}$ & $\mathbf{71.2\pm0.7}$\\
        \bottomrule
        \multicolumn{9}{l}{* The \textsc{g-swm} results are computed with one failed random seed being excluded.}\\
    \end{tabular}
    \end{adjustbox}
\end{table*}
For both tasks, it can be seen that \apex outperforms the baselines by significant margins.
We also observe a smaller standard deviation of the scores for \textsc{apex}, indicating that \textsc{apex} is more stable than the baselines.
We attribute these improvements to the principled mask normalisation, the high-resolution feature maps, and the mask entropy loss. %\hip{we attribute this to...}
Qualitative tracking results for \textsc{apex} are shown in \Cref{fig:teaser_fig}.
It can be seen that \textsc{apex} is clearly able to track individual objects even through occlusions, thus corroborating the quantitative results in \Cref{table:tracking_metrics}.

\begin{figure}
    \centering
    \includegraphics[width=\columnwidth]{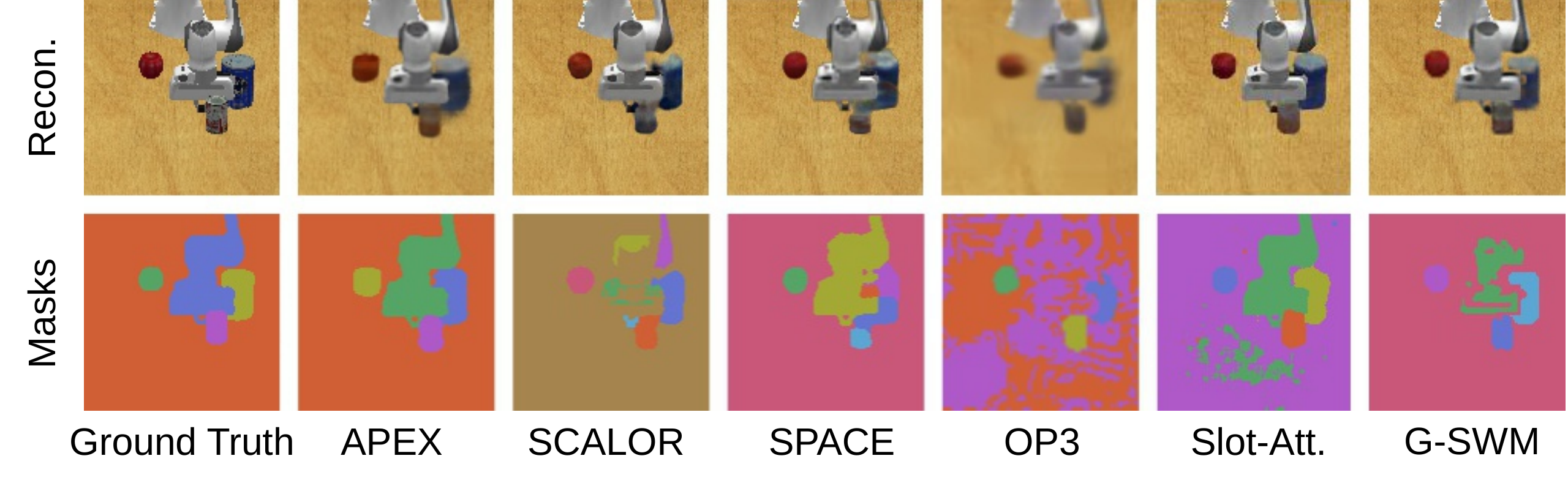}
    \vspace{-0.6cm}
    \caption{Scene segmentation results on P2D. \apex achieves the qualitatively best results, cleanly segmenting all foreground objects as well as the Panda arm.}
    \label{fig:seg_rp}
\end{figure}
\begin{figure}
    \centering
    \includegraphics[width=\columnwidth]{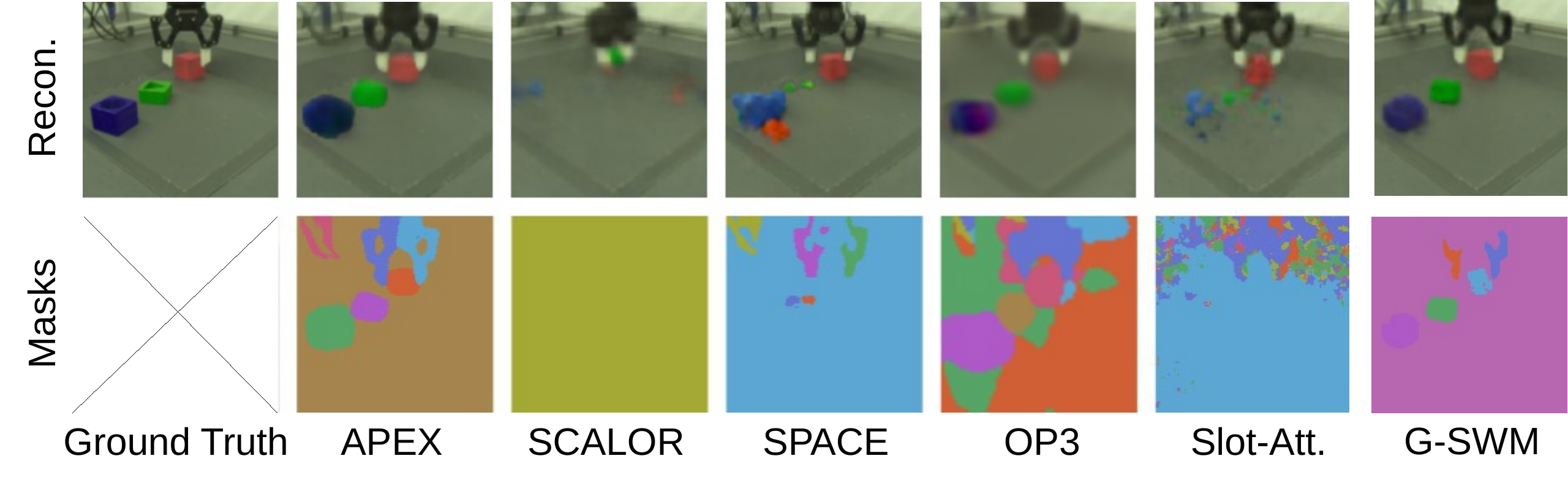}
    \vspace{-0.6cm}
    \caption{Scene segmentation results on Sketchy. In contrast to the baselines, \apex manages to cleanly segment all objects even on this more challenging real-world dataset.}
    \label{fig:seg_sk}
\end{figure}

Qualitative segmentation results for \textsc{apex} as well as for the baselines are shown in \Cref{fig:seg_rp,fig:seg_sk} on P2D and Sketchy, respectively. 
In \Cref{fig:seg_rp}, a fairly cluttered scene from P2D can be seen where the robot arm is interacting with two objects in close proximity to each other. 
While \textsc{apex} clearly segments the foreground objects and the arm itself -- a key prerequisite for task planning and control -- the baselines struggle to do so.
\textsc{scalor} is unable to accurately segment the arm with part of it being captured by the background module.
We conjecture that this is due to the Gaussian image likelihood used for training \textsc{scalor}, as this does not untie the standard deviation of the foreground and the background likelihood as in \textsc{apex}. A similar problem is observed with \textsc{g-swm} which uses the same image likelihood modelling as \textsc{scalor}.
For \textsc{apex} and \textsc{space}, in contrast, a smaller standard deviation is used for the background likelihood ($0.04$) than that of the foreground ($0.1$) to prevent the background module from also capturing the foreground objects.
A Gaussian likelihood with a smaller standard deviation leads to a sharper distribution and thus results in a very low likelihood when the reconstructed colour deviates even slightly from the target.
This will force the background module to focus on the more uniform background pixels which are easier to reconstruct compared to the foreground pixels.
This is further examined in \Cref{abst}.
While \textsc{space} manages to segment the robot arm from the background, it fails to segment the small objects accurately.
Unlike \textsc{apex}, \textsc{space} operates on static images and can thus not leverage temporal information. 
We argue that this helps \apex to learn to distinguish objects even when they are physically very close to each other in one frame, as they might move relative to each other in other frames.
Both \textsc{op3} and \textsc{slot-attention} are able to roughly segment the objects but the segmentation results are noisy and inaccurate.

In \Cref{fig:seg_sk}, \textsc{apex} accurately segments all the objects in the scene -- even the cables of the manipulator.
We attribute the segmentation of the robot manipulator into left and right grippers to the relative motion (open/close) of the two gripper handles.
In contrast, \textsc{scalor} fails to segment the objects and while \textsc{space} is able to segment the arm, it omits the other foreground objects.
This might be caused by the fact that the objects have a uniform colour and are therefore easily reconstructed by the background module.
\textsc{g-swm} successfully segments the objects, but part of the arm is treated as background.
\textsc{op3} and \textsc{slot-attention} struggle to predict accurate masks.

%%%%%%%%%%%%%%%%%%%%%%%%%%%%%%%%%%%%%%%%%%%%%%%%%%%%%%%%%%%%%%%%%%%%%%%%%%%%%%%%
\subsection{Ablation Study}
\label{abst}
A set of ablation experiments is performed to validate the efficacy of the key design choices that set apart \textsc{apex} from prior art: better scene encoding, principled mask normalisation, and a mask entropy loss.
The results are summarised in \Cref{table:ablations}.
\begin{table*}
    \centering
    \caption{Mean and standard deviation of the segmentation metrics on P2D from four random seeds.}
    \label{table:ablations}
    \begin{tabular}{l c c c c}
        \toprule
        & ARI-FG & MSC-FG & ARI & MSC \\
        \midrule
        \apex & $\mathbf{0.89\pm0.02}$ & $\mathbf{0.73\pm0.01}$ & $\mathbf{0.93\pm0.00}$ & $\mathbf{0.80\pm0.01}$ \\
        \midrule
        Image space STN & $0.71\pm0.27$ & $0.62\pm0.10$ & $0.91\pm0.01$ & $0.71\pm0.08$ \\
        No entropy loss & $0.63\pm0.18$ & $0.63\pm0.07$ & $0.92\pm0.01$ & $0.73\pm0.05$ \\
        \textsc{scalor}-norm & $0.62\pm0.30$ & $0.60\pm0.13$ & $0.90\pm0.03$ & $0.70\pm0.09$ \\
        Gaussian likelihood * & $0.10\pm0.01$ & $0.06\pm0.00$ & $0.00\pm0.00$ & $0.28\pm0.00$ \\
        Gaussian likelihood \& \textsc{scalor}-norm. * & $0.02\pm0.00$ & $0.06\pm0.00$ & $0.00\pm0.00$ & $0.28\pm0.00$ \\
        \bottomrule
        \multicolumn{5}{l}{* These models consistently fail to meaningfully segment the images.}\\
    \end{tabular}
\end{table*}
Rather than re-using features extracted by the backbone with STNs when inferring object latents, it can be observed that using STNs to extract information from the input images, such as in \textsc{scalor}, performs consistently worse.
Re-using the backbone features also facilitates a reduction in model parameters and computation. 
The foreground scores when training without the additional mask entropy loss are also smaller, indicating that the entropy loss is indeed beneficial for learning to disambiguate foreground objects.
It can be seen that mimicking the normalisation scheme from \textsc{scalor} results in lower scores compared to the mask normalisation introduced in \Cref{section:apex}.
For the former, the STN outputs a mask $\alpha^s_k \in [0,1]$ for each component and the masks are normalised according to $(\alpha^s_k)^2/\sum_k \alpha^s_k$, which is numerically less stable than a $\operatorname{softmax}$.
This might also explain the increased standard deviation of the scores, especially for the foreground ARI.
Finally, we compare the use of \textsc{apex}'s SGMM image likelihood formulation to using a standard Gaussian likelihood.
Both variants of \textsc{apex} with a standard Gaussian likelihood fail to learn object-centric scene decompositions, highlighting the benefit of the SGMM formulation with separate standard deviations for foreground and background modules.

%%%%%%%%%%%%%%%%%%%%%%%%%%%%%%%%%%%%%%%%%%%%%%%%%%%%%%%%%%%%%%%%%%%%%%%%%%%%%%%%
\subsection{Object Arrangement Task}
\label{sec:arrangement}
Generative models can be used to learn concise and informative representations that can be used in downstream tasks.
In contrast to methods where a single latent vector encapsulates all object-relevant information (e.g. \cite{veerapaneni2020entity,locatello2020object}), further factorising the information into $[\zpres_{k,t}\,, \zwhere_{k,t}\,, \zwhat_{k,t}]$ as well as foreground and background latents allows us to directly use these representations as inputs to a controller.

We demonstrate this in an object arrangement task where a robot is required to pick and place objects on a table according to a \emph{goal image}.
\begin{figure}
    \centering
    \includegraphics[width=1.0\columnwidth]{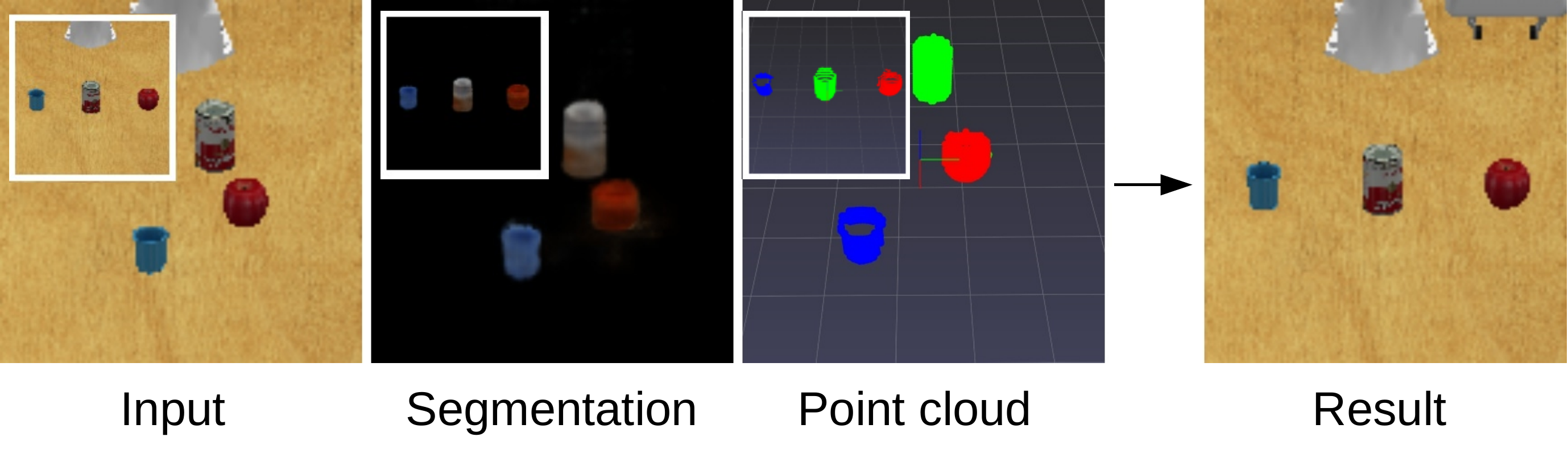}
    \caption{Object arrangement task illustration. Given an input of the current scene and a goal image (a), objects are segmented (b) and matched (c). Depth information is used to obtain the 3D locations and shapes of all objects, which are used as inputs to a heuristic control policy that tries to move the objects to the desired locations specified by the target image.}
    \label{fig:demo_task}
\end{figure}
This image is first parsed into foreground objects and background.
Assuming depth information is available, e.g. via a stereo camera, the 3D location and shape of an object are obtained by filtering a depth image according to the object mask.
The current scene is processed in the same fashion, which allows the current objects to be matched to the associated objects in the goal image according to the smallest $\mathcal{L}_2$ distance in terms of the objects' appearances encoded by $\zwhat_k$.
This is facilitated by discarding \emph{empty detections} according to $\zpres_k$ and the explicit separation of foreground and background components.
The current and desired location and shape of each object serve as inputs to a heuristic control policy that executes a sequence of sub-tasks to move the objects.
This is illustrated in \Cref{fig:demo_task}.
To avoid collisions, a check for whether the target location is occupied by other objects is conducted before executing a sub-task.
If all target locations are occupied, one of the unsorted objects is moved to an edge of the table to make space for the other objects.
This object will be moved to its target space at the end of the re-arrangement process.

Three object arrangement tasks are created involving two to four objects on a table.
For each task, we create $150$ test scenarios with the same group of objects spawned in different locations.
The performance of \textsc{apex} is compared to \textsc{scalor}, \textsc{space} and \textsc{g-swm}.
Comparisons against \textsc{op3} and \textsc{slot-attention} are omitted here as neither explicitly models foreground and background components which are required for task execution.
Performance is quantified using the mean distance of the final object locations relative to the desired object locations.
We set a fixed penalty of $1\mathrm{m}$ for outliers where an object falls off the table.

The results are summarised in \Cref{table:demo_results}.
\begin{table}
    \centering
    \caption{Mean and standard deviation of the object distance  to the desired goal positions in metres.}
    \label{table:demo_results}
    \begin{tabular}{l c c c}
        \toprule
                        & Two objects    & Three objects   & Four objects \\
        \midrule
        \textsc{space}  & $0.07\pm0.16$         & $0.23\pm0.25$         & $0.23\pm0.20$ \\
        \textsc{scalor} & $0.22\pm0.17$         & $0.15\pm0.17$         & $0.18\pm0.21$ \\
        \textsc{g-swm}  & $\mathbf{0.02\pm0.09}$ & $\mathbf{0.04\pm0.12}$         & $0.25\pm0.21$ \\
        \apex           & $0.04\pm0.10$ & $0.06\pm0.13$ & $\mathbf{0.09\pm0.18}$ \\
        \bottomrule
    \end{tabular}
\end{table}
\textsc{apex} performs better than \textsc{space} and \textsc{scalor} on all three tasks.
Some failures are caused by missing detections or incorrect object matches. \textsc{g-swm} performs better than \textsc{APEX} on tasks with two or three objects but fails to segment the goal image properly on task with four objects. Although \textsc{G-SWM} outperforms \textsc{APEX} when there are fewer objects, APEX appears to pull ahead as the number of objects increases. APEX also performs better on segmenting the robot arm (\cref{fig:seg_rp}, \cref{fig:seg_sk}), which is excluded from this task.
\textsc{space} performs worst on tasks with three objects, but better than \textsc{scalor} for two objects. 
We find that \textsc{space} tends to oversegment the objects, i.e., a single object is divided into several components, which reduces the efficacy of both object matching and location estimation.
We hypothesise that the improvement of \textsc{scalor} compared to \textsc{space} is facilitated by the propagation module which is also incorporated into \textsc{apex} and aids the learning of higher quality segmentations.

\vspace{-0.15cm}
\section{Conclusions}

This paper proposes \apex, a novel, object-centric generative model designed to provide state-of-the-art unsupervised object segmentation and tracking on datasets commonly encountered in robotics.
\apex is evaluated on the established Sketchy dataset \cite{cabi2019scaling} for qualitative results and on a custom Panda Pushing Dataset (P2D) for both quantitative and qualitative results.
We show that \textsc{apex} comprehensively outperforms prior art in terms of segmentation and tracking by leveraging improved feature encoding modules as well as a principled normalisation scheme for object and background masks.  Finally, we demonstrate the efficacy of the unsupervised object representations learned by \textsc{apex} on a robot manipulation task that involves the rearrangement of several objects on a table.
\textsc{apex} outperforms most of the baselines due to consistently providing segmentations of significantly higher quality, leading to improvements in object matching and 3D shape extraction.

%\addtolength{\textheight}{+0cm} 
%                                     This command serves to balance the column lengths
%                                   % on the last page of the document manually. It shortens
%                                   % the textheight of the last page by a suitable amount.
%                                   % This command does not take effect until the next page
%                                   % so it should come on the page before the last. Make
%                                   % sure that you do not shorten the textheight too much.

%%%%%%%%%%%%%%%%%%%%%%%%%%%%%%%%%%%%%%%%%%%%%%%%%%%%%%%%%%%%%%%%%%%%%%%%%%%%%%%%

%%%%%%%%%%%%%%%%%%%%%%%%%%%%%%%%%%%%%%%%%%%%%%%%%%%%%%%%%%%%%%%%%%%%%%%%%%%%%%%%

%%%%%%%%%%%%%%%%%%%%%%%%%%%%%%%%%%%%%%%%%%%%%%%%%%%%%%%%%%%%%%%%%%%%%%%%%%%%%%%%
% \section*{APPENDIX}

% Appendixes should appear before the acknowledgment.

\clearpage

\section*{ACKNOWLEDGMENT}
This work was supported by EPSRC Programme Grant (EP/V000748/1), an Amazon Research Award and the China Scholarship Council. The authors would like to acknowledge the use of the University of Oxford Advanced Research Computing (ARC) facility in carrying out this work. http://dx.doi.org/10.5281/zenodo.22558.

%%%%%%%%%%%%%%%%%%%%%%%%%%%%%%%%%%%%%%%%%%%%%%%%%%%%%%%%%%%%%%%%%%%%%%%%%%%%%%%%

\bibliographystyle{IEEEtran}
\bibliography{references}

\end{document}